\documentclass[sigconf,nonacm]{acmart}

\usepackage{amsmath}

\usepackage{amssymb,mathtools}
\usepackage{graphicx}
\usepackage{url}
\usepackage{booktabs}
\usepackage{tabularx}
\usepackage{adjustbox}
\usepackage{microtype}
\title{Online Prediction System for Proactive Autoscaling}

\author{Jun Gao}
\author{Areg Poghosyan}
\author{Pavel Paulau}
\author{Sumit Mohanty}
\author{Tengfei Mu}
\author{Shawn Chen}
\affiliation{
  \institution{Meta Platforms, Inc.}
  \country{USA}
}

\begin{document}

\begin{abstract}
We present an online learning system for short-term time series forecasting that enables proactive autoscaling in large-scale distributed systems.
Unlike traditional forecasting methods that focus on modeling seasonality and long-term trends, the system targets high-accuracy, low-latency predictions for immediate future intervals for dynamic resource management in internet-scale infrastructures.
The system employs a two-stage architecture: a seasonality decomposition stage followed by a residual correction stage using lightweight linear models with engineered features.
By retraining models on-the-fly for every prediction cycle, the online prediction system continuously adapts to changing demand patterns without requiring persistent model storage or feature drift monitoring.
Deployed across many heterogeneous services, the system processes over 900K predictions per hour; in an offline evaluation on production traffic, its STL-based configuration reduces prediction error (MAPE) by 18.6\% relative to the previous production model.
In production, the system has saved an estimated 118K servers daily.
\end{abstract}
\settopmatter{printacmref=false}
\maketitle

\section{Introduction} \label{sec_intro}
\subsection{Motivation}

Modern large-scale internet services are built upon complex, dynamically evolving infrastructures serving billions of users worldwide. One component of managing these systems is \textit{autoscaling}---the process of automatically adjusting computational resources to match fluctuating user demand. The stakes are high: inadequate scaling can lead to catastrophic service outages due to resource exhaustion. 

Traditional autoscaling methods are predominantly reactive, responding to past or current load metrics~\cite{Rzadca2020Autopilot}. This reactive nature creates a fundamental lag: by the time the system detects a traffic spike and provisions additional resources, the spike may have already caused degraded performance or service disruptions. Conversely, when demand drops, reactive systems are slow to release resources, leading to sustained over-provisioning.

Proactive autoscaling based on \textit{time-series forecasting} addresses this limitation by predicting future system load so that resources can be provisioned just-in-time~\cite{Cortez2017,Shahrad2020Serverless}, enhancing both service reliability and efficiency. The core challenge is building a forecasting model capable of handling the scale and heterogeneous nature of data center workloads---from smooth diurnal patterns in social media traffic to highly bursty, event-driven workloads triggered by product launches or breaking news. While prior work has explored scalable distributed training for such models~\cite{Ye2024SHARE}, deploying them in production introduces additional challenges: model maintenance across many services, sub-second prediction latency requirements, and robust generalization across services with vastly different traffic characteristics.

\subsection{Design Requirements}
Beyond forecasting accuracy, a production autoscaling prediction system must satisfy several operational requirements:
\begin{itemize}
\item \textbf{Maintenance.} The system should require minimal human intervention and computational resources, enabling sustainable operation at scale.

\item \textbf{Low latency.} Predictions must be delivered with minimal delay to support real-time autoscaling decisions and timely resource allocation.

\item \textbf{Scalability.} The system must handle various services and prediction requests per day without performance degradation.

\item \textbf{High frequency.} Forecasts should be generated at high temporal resolution (e.g., every 30 minutes) to capture rapid workload changes.

\item \textbf{Explainability.} Outputs should be interpretable to facilitate debugging, operational transparency, and stakeholder trust.
\end{itemize}

\subsection{Existing Solutions}
Prior to the online prediction system, the predominant approach to autoscaler forecasting  relied on polynomial regression models. These systems aggregate throughput data at daily granularity and fit a polynomial curve (typically of fourth degree) to capture trends and seasonality, then extrapolate to forecast future values. Some variants increase temporal resolution by fitting polynomials to hourly data or interpolating daily predictions. While polynomial models are computationally efficient and easy to interpret, their rigid functional form limits their ability to model the complex, non-stationary workload patterns common in production environments.

However, polynomial models exhibit several fundamental limitations in this setting:
\begin{enumerate}
    \item \textbf{Static nature.} Polynomial models are inherently static---once fitted, they do not adapt to sudden workload changes such as traffic spikes from product launches or viral content. This rigidity necessitates large safety buffers (often 20--30\% above predicted demand), reducing efficiency.
    \item \textbf{Limited temporal resolution.} Increasing temporal resolution from daily to hourly does not yield significant accuracy improvements. Polynomial models fundamentally cannot capture short-term fluctuations, abrupt changes, or complex patterns in high-frequency data because they impose a smooth, globally-parameterized curve on inherently non-smooth signals.
    \item \textbf{Overfitting at higher degrees.} Polynomials fitted to noisy hourly data overfit, producing unstable forecasts with oscillatory behavior near the boundaries of the fitting window (Runge's phenomenon). This leads to poor generalization, particularly during periods of atypical demand.
    \item \textbf{No directional awareness.} Polynomial extrapolation provides no mechanism for detecting whether demand is accelerating or decelerating within a prediction window, limiting its usefulness for proactive scaling decisions.
\end{enumerate}

An alternative baseline is trailing prediction ($\hat{y}_t = y_{t-1}$), which uses the most recent observation as the forecast for the next interval. While computationally trivial, this approach is purely reactive and unable to anticipate demand changes. In our experiments (Section~\ref{sec:experiments}), we show that even simple models with proper seasonality decomposition outperform the polynomial baseline.

\subsection{Contributions}
We present a production prediction system designed, deployed, and battle-tested across many services. Our contributions are:
\begin{itemize}
    \item \textbf{Production-scale forecasting system.} We present the architecture and implementation of a real-time, online learning forecasting system for autoscaling, serving many heterogeneous services.

    \item \textbf{Lightweight two-stage modeling.} We introduce a two-stage architecture combining seasonality decomposition (STL~\cite{Cleveland1990STL}) with linear residual correction on engineered features, achieving higher accuracy than the production baseline with sub-second training latency.

    \item \textbf{Stateless online learning.} The system retrains models on-the-fly for every prediction cycle, eliminating the need for model storage, versioning, or feature drift monitoring---key advantages for maintenance at scale.

    \item \textbf{Offline evaluation and production impact.} We compare seven model configurations across two improvement dimensions on production traffic from nine services, and report real-world impact including an estimated 118K servers saved daily.
\end{itemize}

\section{Related Work}
\label{sec:related_work}

Our work is situated at the intersection of time-series forecasting, machine learning for systems (AIOps), and large-scale industrial applications. We review the most relevant literature in these areas.

\subsection{Time-Series Forecasting Models}
Time-series forecasting has a rich history, evolving from classical statistical models like ARIMA~\cite{Box2015} and exponential smoothing to modern deep learning architectures. Classical decomposition methods, notably STL~\cite{Cleveland1990STL}, remain foundational for separating seasonal, trend, and residual components---an approach we adopt in the first stage. Recurrent Neural Networks (RNNs), particularly Long Short-Term Memory (LSTM) networks~\cite{Hochreiter1997}, and their variants like GRU, have been widely adopted for their ability to capture temporal dependencies. Architectures incorporating convolutional layers, such as Temporal Convolutional Networks (TCN)~\cite{Bai2018}, demonstrated the ability to capture long-range patterns efficiently through dilated causal convolutions.

In recent years, Transformer-based models~\cite{Vaswani2017} have achieved remarkable success. Innovations have focused on adapting the self-attention mechanism for temporal data to improve efficiency and performance. For instance, the Informer~\cite{Zhou2021Informer} introduced a \textit{ProbSparse} self-attention mechanism to reduce the quadratic complexity of standard attention, while the Autoformer~\cite{Wu2021Autoformer} proposed a decomposition architecture with an auto-correlation mechanism that explicitly separates trend and seasonal components. More recently, PatchTST~\cite{Nie2023PatchTST} has shown that a simple patching mechanism---segmenting time series into subseries-level patches---can achieve state-of-the-art results with reduced computing. Concurrently, a new line of research has questioned the necessity of complex attention mechanisms for forecasting. Simple linear models like DLinear~\cite{Zeng2023DLinear} and lightweight MLP-based models such as TSMixer~\cite{Chen2023TSMixer} have been shown to outperform complex Transformer models on many benchmarks, sparking a debate about model complexity versus performance. Most recently, state-space models like Mamba~\cite{Gu2023Mamba} are also showing great promise for sequence modeling, offering an alternative with linear complexity. Global neural models such as N-BEATS~\cite{Oreshkin2020NBEATS} learn shared patterns across many series, and pretrained foundation models such as TimesFM~\cite{Das2024TimesFM} and Chronos~\cite{Ansari2024Chronos} forecast previously unseen series zero-shot.

While this vibrant research landscape provides a plethora of powerful models, they are often evaluated on public benchmarks (e.g., ETTh, Weather, Electricity) under controlled conditions with pre-processed, clean data. Our work diverges by focusing on the practical deployment of forecasting models, where challenges such as inference latency (sub-second requirements), model footprint (many concurrent models), robustness to distribution shifts (traffic pattern changes from product launches, incidents, and seasonal events), and generalization across diverse time-series with varying statistical properties are paramount design considerations. Our experimental results (Section~\ref{sec:experiments}) confirm that, in this production setting, simple linear models with proper decomposition outperform CNN-based alternatives, echoing the findings of~\cite{Zeng2023DLinear}.

\subsection{Forecasting for Cloud Resource Management}

The application of machine learning to improve cloud resource management, a field often termed AIOps, is an active area of research. Predictive autoscaling is a key problem in this domain. Early work often relied on simpler models to predict CPU utilization or request counts; Resource Central~\cite{Cortez2017}, for example, learns from historical virtual-machine telemetry to predict utilization and lifetime in a large cloud platform. Seer~\cite{Gan2019Seer} applies deep learning to distributed traces to anticipate quality-of-service violations in cloud microservices.

Prophet, a library developed by Facebook, provides a practical and decomposable time-series model that works well for many business forecasts~\cite{Taylor2018Prophet}, but may not capture the complex, non-linear patterns present in system metrics.
More recent studies have explored learning-based approaches for more accurate predictions. Several works have designed custom forecasting solutions for specific cloud services, such as forecasting database query workloads~\cite{Ma2018QB5000} or provisioning serverless functions from their invocation patterns~\cite{Shahrad2020Serverless}. FIRM~\cite{Qiu2020FIRM} uses reinforcement learning for fine-grained resource management of SLO-oriented microservices. In production, Autopilot~\cite{Rzadca2020Autopilot} automatically sets CPU and memory limits and replica counts for jobs at Google, combining moving-window statistics with machine-learned recommenders.

However, much of the existing academic work focuses on a limited number of services or relies on public, often sanitized, datasets like the Google cluster traces. In contrast, our system is designed from the ground up to be a multi-tenant platform serving many distinct services. It handles diverse statistical properties, from smooth diurnal patterns to highly bursty, event-driven workloads, making generalization a first-class challenge.

\subsection{Large-Scale Forecasting Systems}
Building and maintaining a forecasting system for various time-series presents unique engineering and scientific challenges. Several large tech companies have shared insights into their platforms. Amazon's DeepAR~\cite{Salinas2020DeepAR} was a pioneering work, developing a single RNN-based model that learns global patterns from a large collection of time-series. Uber has reported neural forecasting of extreme events for its marketplace at scale~\cite{Laptev2017Uber}. Similarly, LinkedIn developed Greykite~\cite{Hosseini2022Greykite}, an open-source library designed to provide interpretable and tunable forecasts for business metrics.

These systems provide valuable high-level blueprints and philosophies. Our work contributes to this line of research by providing a detailed case study of a production forecasting system for the specific, high-stakes application of infrastructure autoscaling. We not only describe our modeling choices but also delve into the system architecture for multi-tenancy and the practical lessons learned from operating a system where prediction errors have direct and significant efficiency and reliability implications. Furthermore, we connect our system's design to broader challenges such as scalable distributed training~\cite{Ye2024SHARE}, presenting a holistic view of a modern industrial forecasting platform.

\section{System Architecture}
\label{sec:system}

We present an online prediction system designed for autoscaling forecasting in large-scale distributed systems. Its primary goal is to forecast the immediate future demand for each service entity, enabling proactive resource allocation. The system is integrated with production infrastructure through three main components:

\begin{itemize}
    \item \textbf{Smart Data Aggregator (SDA).} Handles data ingestion, cleaning, and transformation, preparing real-time inputs for the prediction algorithm. The SDA collects throughput metrics from telemetry system, applies outlier filtering and missing value imputation, and aggregates raw signals into 30-minute buckets.
    \item \textbf{Prediction Service.} Processes incoming data, performs feature engineering, trains models online, and generates demand forecasts with associated uncertainty estimates. The service is implemented as a stateless RPC endpoint that can be horizontally scaled independently of the data pipeline.
    \item \textbf{Autoscaling Integration.} Predicted demand values and their standard deviations are fed into the autoscaling system to inform scaling decisions and buffer allocations. The autoscaler uses the predicted demand $\hat{y}_{t+1}$ along with the prediction standard deviation $\sigma$, estimated as the standard deviation of the model's training residuals, to compute a provisioning target of $\hat{y}_{t+1} + \alpha \cdot \sigma$, where $\alpha$ is a configurable safety factor.
\end{itemize}

\begin{figure}[htbp]
  \centering
  \includegraphics[width=\linewidth]{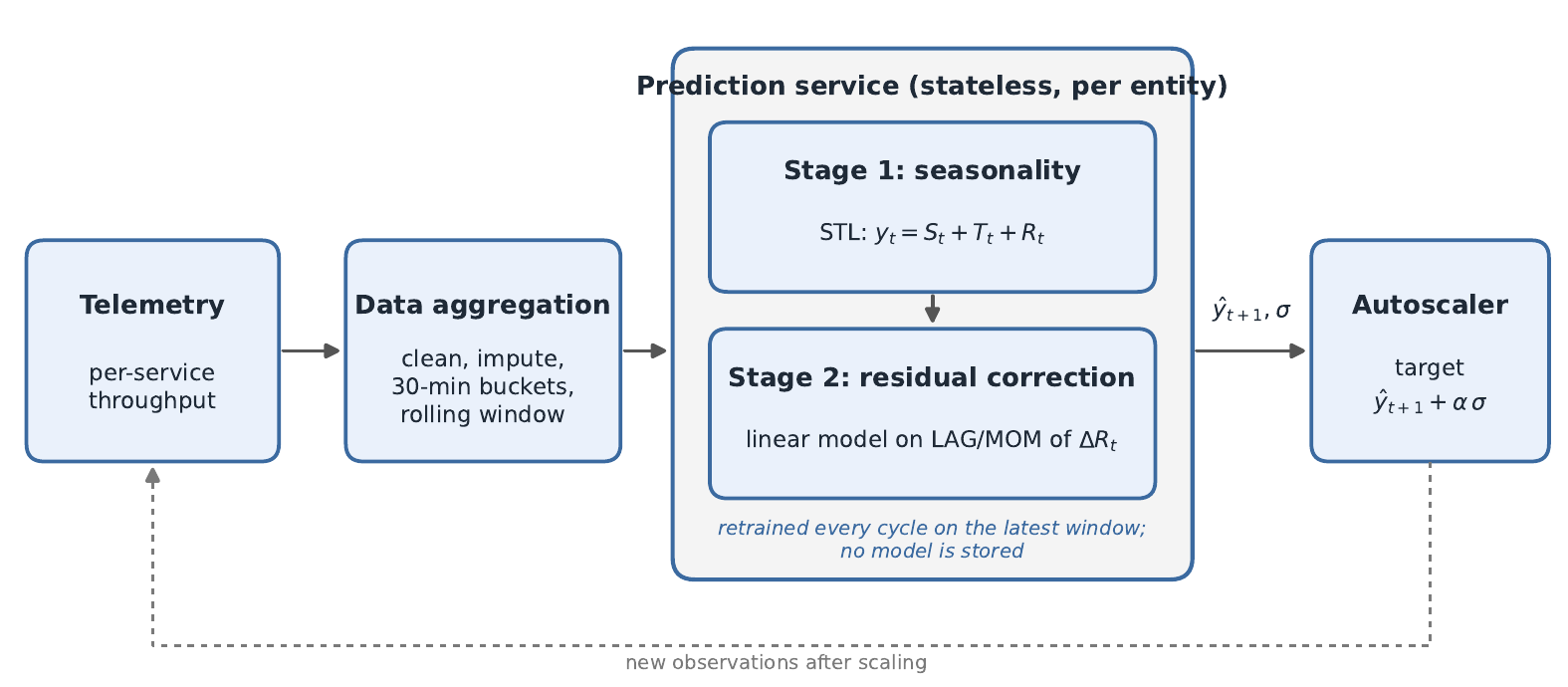}
  \caption{System architecture. The Smart Data Aggregator (SDA) turns raw telemetry into 30-minute buckets over a rolling window. The stateless prediction service fits STL (Stage~1) and a linear residual model (Stage~2) on that window at every cycle and returns $\hat{y}_{t+1}$ and $\sigma$; the autoscaler provisions to $\hat{y}_{t+1}+\alpha\sigma$, and the resulting observations feed the next cycle.}
  \label{fig:architecture}
\end{figure}

\subsubsection{Data Pipeline}
The system ingests streaming throughput data at 30-minute granularity, maintaining a rolling window of recent observations (typically 4 days, or 192 data points) per service entity. This window size was chosen to capture several full daily cycles while keeping the data volume manageable for real-time training. The data pipeline handles several practical challenges: missing data points (due to telemetry failures or service restarts) are interpolated using linear interpolation, and outliers exceeding $3\sigma$ from the rolling mean are clipped to prevent model contamination. Each service entity maintains its own independent data stream, enabling the system to scale horizontally across many entities without cross-entity coordination.

\subsubsection{Feature Engineering}
The prediction algorithm operates on engineered features that capture recent time series dynamics. Table~\ref{tab:features} summarizes the feature families used in the system.

\begin{table}[htbp]
\centering
\caption{Feature families used in  residual correction stage.}
\label{tab:features}
\small
\begin{tabularx}{\columnwidth}{@{}ll>{\raggedright\arraybackslash}X@{}}
\toprule
\textbf{Feature} & \textbf{Lags} & \textbf{Description} \\
\midrule
LAG$_k$ & 10,15,20,25,30 & Raw residual derivative value at lag $k$ \\
MOM$_k$ & 10,15,20,25,30 & Difference between current and lag-$k$ residual derivative \\
\bottomrule
\end{tabularx}
\end{table}

Features are computed on the residual signal (after seasonality removal) and normalized using z-score standardization before model training. The derivative-based features (LAG and MOM on the residual derivative rather than the raw residual) were found to be more informative than level-based features, as they directly capture acceleration and deceleration patterns in demand.

\subsubsection{Online Learning}
For each prediction request, the system retrains a lightweight linear model on the most recent data window. This online learning approach ensures continuous adaptation to changing demand patterns. We deliberately chose simple models (e.g., linear regression) over complex architectures, as the data is inherently noisy and the primary constraint is inference latency rather than model capacity. The complete training-prediction cycle---including data retrieval, feature computation, model fitting, and prediction generation---completes in under 500 milliseconds, enabling the system to process over 900K predictions per hour across all service entities.

\subsubsection{Offline Optimization}
Feature selection and hyperparameter tuning are performed offline through systematic experimentation on held-out validation data spanning multiple weeks of production traffic and diverse service types, so that only the most informative features enter the production pipeline. Key hyperparameters include the data window size, the set of lag values for each feature family, the STL decomposition period, and the smoothing factor applied to predictions; they are optimized via grid search. Other candidate features, such as rates of change and moving averages, added little beyond the LAG and MOM features and were therefore not retained.

\subsection{Two-Stage Prediction Model}
\label{sec:methodology}

The online prediction system employs a two-stage prediction architecture that separates seasonality modeling from residual correction.

\subsubsection{Stage~1: Seasonality Decomposition}
The first stage captures daily periodic patterns in demand. We employ Seasonal-Trend decomposition using Loess (STL)~\cite{Cleveland1990STL} with a period of $P=48$ (corresponding to 24 hours at 30-minute resolution). Given a time series $\{y_1, y_2, \ldots, y_T\}$, STL decomposes each observation as:
\begin{equation}
    y_t = S_t + T_t + R_t,
\end{equation}
where $S_t$ is the seasonal component, $T_t$ is the trend component, and $R_t$ is the residual. STL uses locally-weighted regression (Loess) to iteratively estimate the seasonal and trend components, providing a robust and interpretable seasonality estimate that adapts to changing seasonal patterns without imposing a rigid parametric form. The original production system used polynomial fitting in this stage; following the evaluation in Section~\ref{sec:experiments}, STL replaced it in production. The seasonal forecast for the next time step is obtained by extrapolating the estimated seasonal and trend components.

\subsubsection{Stage~2: Residual Correction}
The second stage models the residual demand $R_t$ not captured by seasonality. Rather than modeling the residual directly, we operate on its first derivative $\Delta R_t = R_t - R_{t-1}$, which is more stationary and easier to predict. We compute derivative features on this signal---lagged values (LAG$_{10,15,20,25,30}$) and momentum indicators (MOM$_{10,15,20,25,30}$)---and train a linear regression model:
\begin{equation}
    \widehat{\Delta R}_{t+1} = \mathbf{w}^\top \mathbf{x}_t + b,
\end{equation}
where $\mathbf{x}_t$ is the feature vector at time $t$ and $\mathbf{w}, b$ are learned parameters. The final prediction combines the seasonal forecast with the residual correction:
\begin{equation}
    \hat{y}_{t+1} = (S_{t+1} + T_{t+1}) + \lambda \cdot (R_t + \widehat{\Delta R}_{t+1}),
    \label{eq:final}
\end{equation}
where $\lambda \in [0, 1]$ is a smoothing factor that controls the weight of the residual correction, balancing responsiveness with stability.

\subsubsection{Online Retraining}
The system retrains models on-the-fly for each prediction cycle. After each prediction, the model is discarded and retrained with updated data for the next cycle. This stateless design eliminates the need for model storage, versioning, or migration infrastructure, and ensures continuous adaptation to evolving demand patterns. The complete cycle---data retrieval, STL decomposition, feature extraction, model training, and prediction---completes within sub-second latency, enabling real-time operation.

\subsubsection{Prediction Target}
The system predicts demand for the next 30-minute interval. The linear model is trained to minimize squared error (RMSE), so it estimates expected demand. Protection against under-provisioning is applied downstream: the autoscaler adds a buffer on top of the forecast (the provisioning target $\hat{y}_{t+1} + \alpha \cdot \sigma$ above). The buffer is tunable per service based on historical traffic characteristics and the service's tolerance for under-provisioning, allowing operators to explicitly control the trade-off between over-provisioning resource consumption and under-provisioning risk.

\subsection{Design Rationale: Simplicity for Speed}
\label{sec:design_rationale}

We deliberately selected simple modeling approaches, prioritizing speed and operational efficiency over maximizing prediction accuracy in isolation. The core motivation was to ensure that the model could be retrained on-the-fly for every prediction request using the most recent data. By aggregating data into 30-minute buckets and using lightweight linear models, both training and inference complete within sub-second latency---enabling the system to retrain immediately after each prediction cycle.

This design choice is grounded in three key observations from our production experience:
\begin{enumerate}
    \item \textbf{Data scarcity per entity.} Each service entity has only $\sim$192 data points (4 days at 30-minute granularity) available for training. Complex models with many parameters tend to overfit on such limited data, as confirmed by our CNN experiments (Section~\ref{sec:experiments}).
    \item \textbf{Non-stationarity.} Production workloads are inherently non-stationary, with traffic patterns shifting due to product changes, seasonal events, and infrastructure migrations. A model trained on last week's data may not generalize to this week's traffic. Online retraining with fresh data inherently handles this challenge.
    \item \textbf{Operational issues.} Deploying and maintaining complex models (e.g., deep neural networks) across many services would require GPU infrastructure, model versioning, monitoring for training failures, and specialized ML engineering support. Linear models eliminate all of these issues.
\end{enumerate}

In our setting, this choice does not sacrifice accuracy: in our experiments (Section~\ref{sec:experiments}), the linear STL+LR(LAG/MOM) model is more accurate than every CNN-based alternative, while also being more reliable, easier to maintain, and simpler to deploy.

\section{Experiments}
\label{sec:experiments}

We conduct a systematic evaluation of the framework, comparing multiple modeling strategies across two improvement dimensions: (A) the seasonality decomposition method in Stage~1 (polynomial fitting vs.\ STL), and (B) the residual feature extraction method in Stage~2 (hand-crafted LAG/MOM features vs.\ learned CNN features). We evaluate seven model configurations on real production traffic data from autoscaling infrastructure.

\subsection{Experimental Setup}

\subsubsection{Data}
We sample 9 representative service entities from production autoscaler fleet via the \texttt{autoscaler\_\allowbreak prediction\_\allowbreak entities\_\allowbreak ods} table, spanning diverse traffic patterns including smooth diurnal curves, bursty event-driven workloads, and gradually trending services. For each entity, we retrieve 4 days (192 data points at 30-minute granularity) of throughput time series from the ODS telemetry system. The data is split into 80\% training and 20\% test sets, preserving temporal order.

\subsubsection{Models}
We compare seven model configurations organized along two dimensions:

\begin{itemize}
    \item \textbf{Stage~1 (Seasonality):}
    \begin{itemize}
        \item \textit{Polyfit}: 4th-degree polynomial fitting on daily-aggregated throughput, the production baseline at the time of this evaluation.
        \item \textit{STL}: Seasonal-Trend decomposition using Loess~\cite{Cleveland1990STL} with period $P=48$ (24 hours at 30-minute resolution), extracting seasonal and trend components for more robust seasonality capture.
    \end{itemize}
    \item \textbf{Stage~2 (Residual Correction):}
    \begin{itemize}
        \item \textit{None}: Seasonality-only baseline with no residual correction.
        \item \textit{LR(LAG/MOM)}: Linear regression on 10 hand-crafted features---lagged values (LAG$_{10,15,20,25,30}$) and momentum indicators (MOM$_{10,15,20,25,30}$)---operating on the residual derivative signal. This was the production approach at the time of this evaluation.
        \item \textit{CNN features}: A 2-layer CNN trained to predict residual derivatives from a sliding window of 30 raw values (15 hours). CNN$\rightarrow$LR denotes CNN-extracted features fed to a scikit-learn \texttt{LinearRegression}.
    \end{itemize}
\end{itemize}

Table~\ref{tab:models} summarizes the full set of seven configurations.
\begin{table}[htbp]
\centering
\caption{Model configurations evaluated in the experiment.}
\label{tab:models}
\small
\setlength{\tabcolsep}{4pt}
\begin{tabularx}{\columnwidth}{@{}cll>{\raggedright\arraybackslash}X@{}}
\toprule
\textbf{\#} & \textbf{Model} & \textbf{Stage~1} & \textbf{Stage~2} \\
\midrule
1 & Polyfit\_Only           & Polyfit & --- \\
2 & Polyfit+LR(LAG/MOM)    & Polyfit & LR(LAG/MOM) \\
3 & Polyfit+CNN            & Polyfit & CNN (2-layer) $\rightarrow$ linear head \\
4 & Polyfit+CNN$\rightarrow$LR & Polyfit & CNN features $\rightarrow$ sklearn LR \\
5 & STL\_Only               & STL     & --- \\
6 & STL+LR(LAG/MOM)        & STL     & LR(LAG/MOM) \\
7 & STL+CNN$\rightarrow$LR  & STL     & CNN features $\rightarrow$ sklearn LR \\
\bottomrule
\end{tabularx}
\end{table}

\subsubsection{Metrics}
We evaluate models using three complementary metrics:
\begin{itemize}
    \item \textbf{MAPE} (Mean Absolute Percentage Error): Primary accuracy metric, measuring relative prediction error. Lower is better.
    \item \textbf{Normalized RMSE}: Root mean squared error normalized by the mean of actual values, penalizing large deviations. Lower is better.
    \item \textbf{Directional Accuracy}: Fraction of predictions that correctly capture the direction of throughput change (increase vs.\ decrease). Higher is better.
\end{itemize}
Underprediction rate depends on the buffer configured in production, so we treat it as an online metric and discuss it with the deployment results in Section~\ref{sec:impact}.

\subsection{Results}

Table~\ref{tab:results} presents the aggregated results across all 9 entities, sorted by MAPE.

\begin{table}[htbp]
\centering
\caption{Aggregated experiment results across 9 production entities. Best results per column are in \textbf{bold}. $\Delta$MAPE is the relative change versus the production baseline, Polyfit+LR(LAG/MOM) (negative is better).}
\label{tab:results}
\small
\begin{adjustbox}{max width=\columnwidth}
\begin{tabular}{@{}lcccc@{}}
\toprule
\textbf{Model} & \textbf{MAPE} & \textbf{Norm RMSE} & \textbf{Dir. Acc.} & $\Delta$\textbf{MAPE} \\
\midrule
STL+LR(LAG/MOM)    & \textbf{6.68\%} & \textbf{8.01\%} & 69.9\% & $-$18.6\%  \\
STL+CNN$\rightarrow$LR & 7.23\%          & 8.35\%          & 69.1\% & $-$12.0\%  \\
Polyfit+CNN            & 7.53\%          & 9.15\%          & 52.0\% & $-$8.3\%   \\
Polyfit+CNN$\rightarrow$LR & 7.71\%      & 9.35\%          & 50.1\% & $-$6.1\%   \\
Polyfit+LR(LAG/MOM)    & 8.21\%          & 10.02\%         & 49.1\% & ---        \\
STL\_Only               & 8.30\%          & 9.65\%          & \textbf{71.8\%} & $+$1.1\%   \\
Polyfit\_Only           & 9.97\%          & 11.95\%         & 59.1\% & $+$21.4\%  \\
\bottomrule
\end{tabular}
\end{adjustbox}
\end{table}

\subsection{Analysis}

\subsubsection{STL dramatically improves seasonality modeling}
Replacing polynomial fitting with STL decomposition yields the single largest improvement across all configurations. STL\_Only (8.30\% MAPE) nearly matches the production baseline, Polyfit+LR(LAG/MOM) (8.21\%), despite using no residual correction at all. When combined with the same LAG/MOM features used in production, STL+LR(LAG/MOM) achieves 6.68\% MAPE---an 18.6\% relative reduction over the production baseline. This improvement is attributable to STL's ability to adaptively decompose the time series into seasonal and trend components without imposing a rigid polynomial form, making it more robust to non-stationary traffic patterns.

Notably, STL-based models achieve dramatically higher directional accuracy ($\sim$70\% vs.\ $\sim$50\% for polyfit-based models), indicating that STL better captures the direction of short-term traffic fluctuations. This property is particularly valuable for autoscaling, where correctly anticipating the direction of demand change is often more important than minimizing absolute error.

\subsubsection{CNN features provide moderate improvements}
Replacing hand-crafted LAG/MOM features with CNN-extracted features improves polyfit-based models from 8.21\% to 7.53\% MAPE (8.3\% relative improvement).

The CNN$\rightarrow$LR variant, which uses CNN as a feature extractor feeding into scikit-learn \texttt{LinearRegression}, slightly underperforms the direct CNN prediction head on polyfit models (7.71\% vs.\ 7.53\% MAPE), so fitting a linear model on CNN-extracted features brings no additional benefit in this setting.

\subsubsection{STL subsumes CNN's temporal learning}
An important finding is that CNN features provide much less incremental benefit when combined with STL than with polyfit. STL+CNN$\rightarrow$LR (7.23\%) underperforms STL+LR(LAG/MOM) (6.68\%), indicating that STL already captures much of the temporal structure that CNNs learn from raw data. The hand-crafted LAG/MOM features, designed to capture momentum and recent changes, complement STL's seasonality decomposition more effectively than CNN features do.

\subsubsection{Practical implications}
The most impactful upgrade path was also the simplest, and it is the one we deployed in production: replacing polynomial fitting with STL in Stage~1 while retaining the existing LAG/MOM features in Stage~2. This requires no neural network inference, no GPU resources, and no changes to the feature engineering pipeline---only a swap of the seasonality component. The STL library (\texttt{statsmodels.\allowbreak tsa.\allowbreak seasonal}) is well-maintained, computationally lightweight, and introduces negligible additional latency for online retraining.

\subsection{Discussion}

Our experiments reveal a clear hierarchy of improvements: seasonality modeling (Stage~1) matters more than residual feature engineering (Stage~2) for the online prediction system's accuracy. This finding aligns with the time-series forecasting literature, where proper decomposition of seasonal patterns is often the single most impactful modeling choice~\cite{Cleveland1990STL}. The result also validates the two-stage architecture: by cleanly separating seasonality from residual correction, each component can be independently improved and evaluated.

Table~\ref{tab:comparison} provides a qualitative comparison  with alternative forecasting approaches considered during the system's development.

\begin{table}[htbp]
\centering
\caption{Qualitative comparison  with alternative approaches.}
\label{tab:comparison}
\small
\begin{adjustbox}{max width=\columnwidth}
\begin{tabular}{@{}lccccc@{}}
\toprule
\textbf{Approach} & \textbf{Accuracy} & \textbf{Latency} & \textbf{Maint.} & \textbf{Adapt.} & \textbf{Interp.} \\
\midrule
Trailing ($\hat{y}_t = y_{t-1}$) & Low & \checkmark & \checkmark & \checkmark & \checkmark \\
Polynomial fit & Medium & \checkmark & \checkmark & $\times$ & \checkmark \\
ARIMA/Prophet & Medium & $\sim$ & $\sim$ & $\times$ & \checkmark \\
LSTM/TCN & High & $\times$ & $\times$ & $\sim$ & $\times$ \\
Transformer & High & $\times$ & $\times$ & $\sim$ & $\times$ \\
\textbf{Ours (STL+LR)} & \textbf{High} & \checkmark & \checkmark & \checkmark & \checkmark \\
\bottomrule
\end{tabular}
\end{adjustbox}
\end{table}

The strong performance of STL+LR(LAG/MOM)---using only linear models and simple feature engineering---further supports our design principle of preferring lightweight, interpretable models for production deployment (Section~\ref{sec:design_rationale}). While deep learning approaches show promise, their marginal accuracy improvements (at most 8.3\% relative improvement from CNN features vs.\ 18.6\% from STL) do not justify the additional complexity, latency, and maintenance need in a system where models are retrained online for every prediction cycle across many services. This finding echoes recent work by~\cite{Zeng2023DLinear}, which demonstrated that simple linear models can match or exceed Transformer-based models on standard forecasting benchmarks.

\section{Real-World Impact}
\label{sec:impact}

The online prediction system has been deployed in production for many years, delivering substantial improvements in both cost efficiency and operational reliability.

\subsection{Operational Improvements}
Across autoscaled services, the online prediction system consistently delivers 3.6--4.3\% efficiency improvements over the previous polynomial-based models. Table~\ref{tab:savings} summarizes the key metrics.

\begin{table}[htbp]
\centering
\caption{Production impact of deployment.}
\label{tab:savings}
\begin{tabular}{@{}lc@{}}
\toprule
\textbf{Metric} & \textbf{Value} \\
\midrule
Daily servers saved & 118K servers \\
WWW service savings & 18--19K servers \\
Prediction throughput & $>$900K predictions/hour \\
\bottomrule
\end{tabular}
\end{table}

\subsection{Web Service Deployment}
The rollout to web services was a major milestone that required overcoming significant technical challenges. WWW comprises more than 500 job handles, each representing a distinct workload with its own traffic pattern. Processing all handles sequentially took more than 5 minutes per prediction cycle, well beyond the latency budget. Because predictions for different job handles are independent, we redesigned the service to run them in parallel, which reduced end-to-end prediction latency for the entire WWW service to under 30 seconds.

The deployment followed an incremental rollout strategy: predictions were initially run in shadow mode alongside the existing polynomial model, with results compared daily. After two weeks of shadow testing confirmed consistent improvements, the system was gradually rolled out to 10\%, 50\%, and finally 100\% of WWW’s traffic. The result was a 3.6\% efficiency improvement, saving 18--19K servers.

\subsection{Reliability}
The online prediction system improved prediction accuracy and adaptability to throughput changes, enabling proactive autoscaling and better resource utilization. Both the frequency of under-estimation events and their average magnitude were reduced compared to the polynomial baseline.

Notably, the system’s online retraining capability allows it to adjust to demand shifts faster than batch-trained alternatives. When a traffic pattern changes---for example, due to a product launch or a shift in user behavior---it adapts within 2--3 prediction cycles (1--1.5 hours), compared to the 24--48 hours required for the polynomial model to be re-fitted. This rapid adaptation is particularly helpful during traffic transitions such as peak-to-off-peak and vice versa, where the polynomial model’s slow response frequently triggered either over-provisioning or capacity alerts.

\subsection{Operational Efficiency}
The stateless, online learning design provides several operational advantages over traditional ML-based forecasting systems:
\begin{itemize}
    \item \textbf{No model storage.} Models are trained and used in-memory, eliminating the need for model versioning, backup, or migration infrastructure. This removes an entire class of operational failures related to model corruption, version mismatch, or storage system outages.
    \item \textbf{No drift monitoring.} Continuous online retraining inherently adapts to distribution shifts, removing the need for dedicated drift detection mechanisms. In contrast, batch-trained models require active monitoring and periodic retraining pipelines.
    \item \textbf{Horizontal scalability.} The stateless architecture enables trivial horizontal scaling---new instances require no initialization, warm-up, or synchronization. During peak periods, additional instances can be spun up in seconds.
    \item \textbf{Minimal human intervention.} Model updates are automatic and continuous, requiring no manual retraining pipelines, specialized ML engineering support, or on-call rotation for model maintenance.
\end{itemize}

\section{Lessons Learned}
\label{sec:lessons}

Deployment at production scale revealed several important lessons:

\paragraph{Balancing reactivity and stability.}
The model’s high reactivity---while beneficial for tracking rapid demand increases---occasionally resulted in overly aggressive downsizing during off-peak periods, increasing the risk of capacity shortfalls when demand rebounded. This led to the introduction of a smoothing factor $\lambda$ (Equation~\ref{eq:final}) and constraints on the maximum downscale rate per prediction cycle. In practice, we found that limiting downscale to at most 5\% per cycle provided a good balance between responsiveness and operational safety.

\paragraph{Feature validation.}
Adding new features (e.g., drain event counters, day-of-week indicators) sometimes inadvertently increased underestimation frequency in certain service types. For example, a drain event feature that improved predictions for stateful services caused overfitting for stateless services with no drain events. This highlighted the need for rigorous A/B validation of all model changes before full rollout, including stratified evaluation across different service categories.

\paragraph{Automation of monitoring.}
Early deployments required significant manual effort to validate model performance---engineers manually inspected prediction plots for each service after every model change. This was unsustainable as the system scaled to many services. Automating evaluation and alerting became a priority, leading to the development of standardized dashboards, automated shadow testing pipelines, and alerting on key metrics (MAPE regression, underprediction rate increase, model staleness).

\paragraph{Stakeholder trust and adoption.}
Service owners reported increased trust in autoscaling due to improved accuracy and transparency. However, they also requested more granular control and visibility. Incremental rollout strategies---deploying to lower tiered services first, then gradually expanding to higher tiered services---proved essential for building confidence. Close collaboration between data science, engineering, and service owner teams was essential throughout the deployment process, particularly in defining service-specific parameters such as buffer sizes and maximum downscale rates.

\section{Conclusion}
\label{sec:conclusion}

We have presented an online prediction system, a scalable online learning system for proactive autoscaling forecasting deployed across many services. Through a two-stage architecture combining STL seasonality decomposition with linear residual correction, the STL-based configuration reduces MAPE by 18.6\% relative to the previous production model in offline evaluation, while maintaining sub-second training latency. Our systematic evaluation across seven model configurations demonstrates that proper seasonality decomposition (Stage~1) yields larger accuracy improvements than sophisticated feature learning (Stage~2), and that simple linear models with engineered features outperform CNN-based alternatives in this production setting.

The system's stateless, online learning design eliminates the need for model storage, versioning, and drift monitoring, enabling efficient operation at scale. In production, it has saved an estimated 118K servers daily, while processing over 900K predictions per hour. The lessons learned from this deployment---including the importance of balancing reactivity with stability, the need for automated monitoring at scale, and the value of incremental rollout strategies---provide practical insights for building and operating forecasting systems in large-scale, high-stakes infrastructure environments.

\section{Future Work}
\label{sec:future}

Several directions remain for future exploration:

\paragraph{Joint demand-supply modeling.}
The current system models only the demand side. A natural extension is to jointly predict available supply (compute resources, network bandwidth), enabling the autoscaler to make more holistic decisions. Techniques such as multi-agent reinforcement learning could capture the dynamic interplay between demand and supply, accounting for planned maintenance windows, hardware failures, and cross-service resource contention.

\paragraph{Automated parameter tuning.}
While initial fine-tuning of model parameters and buffer sizes has yielded improvements, automated hyperparameter optimization (e.g., Bayesian optimization or contextual bandits) could adaptively tune the smoothing factor $\lambda$, buffer sizes, and feature lag sets in real time across diverse workload conditions. This would reduce the need for manual per-service configuration while potentially discovering service-specific parameter settings that improve accuracy.

\paragraph{External signals and anomaly robustness.}
Integrating external data sources---such as calendar events, product launch schedules, marketing campaigns, and regional holidays---could improve predictions for rare or bursty events that are difficult to forecast from historical traffic alone. Complementary anomaly detection techniques would further enhance robustness to unexpected demand spikes, enabling the system to distinguish between genuine demand increases and transient anomalies.

\paragraph{Uncertainty quantification.}
Developing calibrated prediction intervals---beyond the current point predictions with standard deviation estimates---would enable risk-aware autoscaling decisions. Conformal prediction~\cite{Vovk2005} offers a promising framework for distribution-free prediction intervals that could be integrated into the online learning pipeline with minimal computational consumption.

\section*{Acknowledgements}
We thank our colleagues at Meta for their invaluable support and collaboration throughout the development and deployment.

\section{Appendix}
\bibliographystyle{ACM-Reference-Format}
\bibliography{ref}

\end{document}